\documentclass[11pt,conference,letterpaper]{IEEEtran}
\IEEEoverridecommandlockouts

\ifCLASSINFOpdf
  \usepackage[pdftex]{graphicx}
  \graphicspath{{../figures/}}
  \DeclareGraphicsExtensions{.png}
\else
\fi

\usepackage{amsmath}

\usepackage{algorithmicx}
\usepackage{algorithm}

\RequirePackage[
        defernumbers=true,
        backend=biber,
        bibencoding=utf8,
        natbib=true,
        hyperref=false,
        backref=false,
        urldate=long,
        style=numeric,
        citestyle=numeric,
        sorting=none,
        date=year
    ]{biblatex}
\makeatletter
\let\blx@rerun@biber\relax
\makeatother

\begin{document}
\title{Reinforcement Learning Algorithm\\for Traffic Steering in Heterogeneous Network}

\author{\IEEEauthorblockN{Cezary Adamczyk}
\IEEEauthorblockA{Institute of Radiocommunications\\
Poznan University of Technology\\
Poznan, Poland\\
Email: cezary.adamczyk@student.put.poznan.pl}
\and
\IEEEauthorblockN{Adrian Kliks}
\IEEEauthorblockA{Institute of Radiocommunications\\
Poznan University of Technology\\
Poznan, Poland\\
Email: adrian.kliks@put.poznan.pl}}

% conference papers do not typically use \thanks and this command
% is locked out in conference mode. If really needed, such as for
% the acknowledgment of grants, issue a \IEEEoverridecommandlockouts
% after \documentclass

% make the title area
\maketitle

\begin{abstract}
Heterogeneous radio access networks require efficient traffic steering methods to reach near-optimal results in order to maximize network capacity. This paper aims to propose a novel traffic steering algorithm for usage in HetNets, which utilizes a reinforcement learning algorithm in combination with an artificial neural network to maximize total user satisfaction in the simulated cellular network. The novel algorithm was compared with two reference algorithms using network simulation results. The results prove that the novel algorithm provides noticeably better efficiency in comparison with reference algorithms, especially in terms of the number of served users with limited frequency resources of the radio access network.\footnote{Copyright (c) 2021 IEEE. Personal use is permitted. For any other purposes, permission must be obtained from the IEEE by emailing pubs-permissions@ieee.org. This is the author's version of an article that has been published in this journal. Changes were made to this version by the publisher prior to publication. The final version of record is available at https://doi.org/10.1109/WiMob52687.2021.9606426}
\end{abstract}


\IEEEpeerreviewmaketitle

\section{Introduction}
Evolution of telecommunication technologies allows cellular network operators to use radio resources more efficiently and therefore improve service quality and serve more users. Introduction of each new generation of radio access network architecture and gradual implementation of the new technology leads to a situation where several radio access technologies coexist creating a heterogeneous network (HetNet) \cite{Agiwal2016}. Additional heterogeneity in network architecture is caused by many base stations types being installed for different coverage scenarios. These include macro cells for large area coverage and small cells (i.e. micro-, pico-, femtocells) for more throughput-demanding coverage or areas not covered by macro cells \cite{rrm_hetnets}.
\par
Such complex and diverse RAN architecture causes many challenges for the network operator, including backhaul provision for each base station, inter-RAT and intra-RAT interference mitigation and fine-tuning base station parameters to maximize its capacity \cite{Liu2016}. One of many important aspects of HetNets optimization is traffic steering, i.e. load balancing \cite{Khaled2021}. It is the process of adaptive traffic allocation to different base stations (or in more generic approach - available radio resources), often with a certain priority set by the network operator, e.g. to maximize total network throughput.
\par
As in heterogeneous networks (HetNets) one user can often be in range of many base stations, the load balancing algorithm must decide on which cell's radio resources should be allocated to the user. Basic traffic steering algorithms rely on a single criterion to make a decision on what cell should serve given user. This criterion may be current load percentages of in-range base stations or satisfaction of user's bit rate demand with radio resources available to be allocated by each base station. These approaches guarantee moderate efficiency at a low computational cost, but are far from optimal.
%An example of such scenario is when a user located at the edge of a lightly-loaded macro cell demands high bit rate and is also located near a highly-loaded micro cell. Because the user is far away from the macro cell's antennas, the user can not be efficiently served by the cell, because of poor radio link quality (and therefore low reported CQI index). If the macro cell would be to serve the user, it would need to allocate a large portion of its radio resources to satisfy the user's demand for bit rate. This might leave many of its future users not served because of insufficient resources. In such situation, allocating the user to the micro cell instead would be optimal, but an algorithm based on cell load only would decide to serve the user inefficiently with macro cell's resources.
\par
The key paper contribution of this paper is a proposal of an universal traffic steering algorithm for utilization in HetNets. The solution includes a dedicated data-processing flow that combines ANN inference and SARSA algorithm for ANN optimization to provide near-optimal traffic steering. The idea itself is inspired by work described in \cite{rl_jrrm}, but aims to provide more universal utility for all generations of radio access network, including 5G and beyond.

\section{Traffic steering methods}
\subsection{Problem statement and reference scenario}
The key problem addressed in this paper deals with traffic steering in HetNet scenarios. In general, the goal of traffic steering algorithms is to serve users using radio access network's resources in a way that uses the resources most optimally in regard to a chosen criterion. Basic approaches mentioned in \cite{rl_jrrm} include allocating radio resources of the least loaded cell (denoted hereafter as Classic Load Balancing, \emph{CLB}) or allocating radio resources of the cell that provides best user satisfaction (Satisfaction-based Load Balancing, \emph{SLB}). In our approach, we aim to assign users to the cells in order to maximize total user satisfaction.

\subsection{Reinforcement Learning Load Balancing}
In the proposed scheme, we steer the cellular traffic within the HetNet by utilizing a reinforcement learning scheme called SARSA for optimization of an ANN \cite{rl_sutton}. In turn, the ANN adapts its model to use the limited radio resources in a way that maximizes the total user satisfaction in given scenario. The name of the SARSA method comes from the components used to update the state-action value function estimate \cite{rl_sutton}:
\begin{itemize}
    \item \( S_t \) - environment state observation in \textit{t}
    \item \( A_t \) - action taken in \textit{t}
    \item \( R_{t+1} \) - reward received in \textit{t}+1 after taking action \( A_t \)
    \item \( S_{t+1} \) - environment state observation in \textit{t}+1
    \item \( A_{t+1} \) - action taken in \textit{t}+1.
\end{itemize}
The update step applied in SARSA is defined as in (\ref{eq:sarsa_update}) \cite{rl_sutton}. One can observe that, besides the components listed above, it utilizes a learning factor \(\alpha\), that determines how much the new data influences the current estimate, and a discount factor \(\gamma\) to discount influence from predicted state on the current estimate.
\begin{gather}
    Q(S_t, A_t) \xleftarrow{} (1 - \alpha)Q(S_t, A_t) +\nonumber\\
    + \alpha(R_{t+1} + \gamma Q(S_{t+1}, A_{t+1}))
\label{eq:sarsa_update}
\end{gather}
\par
As shown in \cite{rl_sutton}, the probability of the SARSA algorithm convergence to optimal policy of the agent is 100\% if the \emph{\(\epsilon\)-greedy} policy is utilized. It means that for each environment state observation, the action taken by the agent is random with a probability of \(\epsilon\), otherwise it results from the agent's policy. This approach has been utilized in the proposed algorithm.
\par
In the context of SARSA application for traffic steering in HetNets, the environment is constituted by the considered radio access network area together with its active users. Next, the observation of the environment is a set of decision criteria for a single user. The reward after performing an action by the agent is the user's satisfaction after being served. In our approach, the ANN works as the agent, and its policy is modified according to the reward received from the environment after each action taken. This means that the ANN's connection weights and biases are modified according to (\ref{eq:sarsa_update}).
\par
The probability of a random action \(\epsilon\) resulting from the adopted policy of the \emph{\(\epsilon\)-greedy} agent has been implemented in such a way that with successive simulation episodes it decreases by a certain \(\epsilon_{dec}\) value, descending to zero. Such a mechanism is aimed at reducing the impact of random decisions, when the optimization of the agent's policy allows obtaining satisfactory results. Based on the trials of the RLLB method with various combinations of the SARSA method parameters, the following values of the parameters were decided: $\epsilon = 0.1$, $\epsilon_{dec}=10^{-5}$, $\alpha=0,15$, and $\gamma=0,95$. 
% parameter variant presented in Table \ref{tab:sarsa_params} was decided.
% \begin{table}[!hbt]
% \renewcommand{\arraystretch}{1.3}
% \caption{SARSA method parameters utilized in the RLLB method}
% \label{tab:sarsa_params}
% \centering
% \begin{tabular}{|c|c|c|c|}
%      \hline
%      \(\epsilon\) & \(\epsilon_{dec}\) & \(\alpha\) & \(\gamma\) \\
%      \hline
%      0,1 & \(10^{-5}\) & 0,15 & 0,95 \\
%      \hline
% \end{tabular}
% \end{table}
\par
In order for the ANN to be able to optimize its behaviour an enhanced set of criteria is processed. In particular, for each user-cell pair, a set of three parameters is used: current cell load, percentage of cell's available radio resources that would be used up upon serving the user, and an estimated remaining number of users to be handled by the cell. The last parameter may be difficult to determine, however, in a real scenario, the initial process of load balancing could utilize one of the reference algorithms. Then, the number of users handled by the reference method can be used as input for the RLLB method.
\begin{figure*}[!tb]
\centering
\vspace{0.03in}
\includegraphics[width=\textwidth,angle=0]{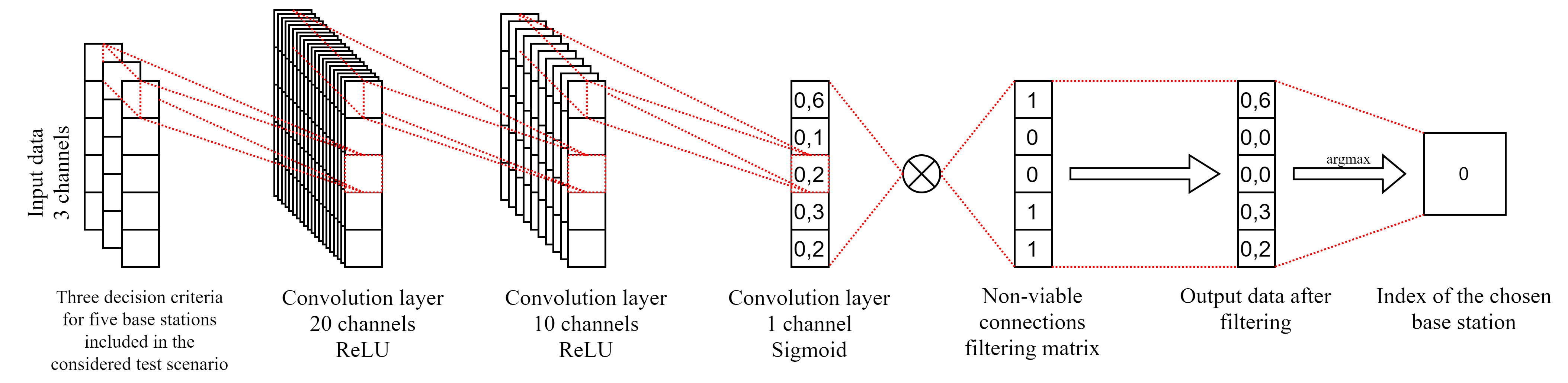}
\caption{Artificial neural network structure used in RLLB method}
\label{fig:rllb_cnn}
\end{figure*}
\par
Regarding the ANN structure, the selection of the number of convolution layers and the number of channels in each layer was performed on the basis of the method efficiency tests for various layer structures. For a large number of layers or channels in each layer, the efficiency of the RLLB method did not improve in relation to a network with a less complex structure. With too few layers or channels in each layer, the RLLB method lead to random decisions and its performance did not improve with subsequent simulation episodes. Finally, a decision was made on the structure of a neural network with three convolution layers with the number of channels 20, 10 and 1. Additionally, the ReLU activation function (i.e. a function that outputs the input directly if it is positive; otherwise, it outputs zero) was used in the first two layers, which significantly improved the stability of the network. Next, the Sigmoid activation function (i.e. a function transforming input data to values from 0 to 1 \cite{neural_networks_haykin}) was used in the third convolution layer, as it allows to obtain values that are convenient for interpretation. Thus, the output values of the neural network can be treated as an assessment of the goodness of the allocation of resources of a given base station to the user expressed as a percentage. Final structure of the neural network utilized in the RLLB method is shown in Figure \ref{fig:rllb_cnn}.
%\vspace{0.1cm}
%Additionally, the ReLU activation function was used in the first two layers, which significantly improved the stability of the network, and next, the Sigmoid activation function was used in the third convolution layer, as it allows to obtain values that are convenient for interpretation \cite{neural_networks_haykin}.

Values at the output of the last convolution layer are subject to additional filtering to exclude base stations with insufficient signal coverage in the position of the currently considered user (i.e. channel quality indicator or user satisfaction equal to 0). The algorithm handles the user using resources of the base station with index equal to the index of the highest value after filtering the output of the ANN. % procedure of the RLLB method is shown in Algorithm \ref{alg:rllb}.

% \begin{algorithm}[!t]
%     \caption{Traffic steering procedure in the RLLB method}
%     \label{alg:rllb}
%     \textbf{Input data:}
%     \begin{itemize}
%         \item \emph{cells} - list of cells in the considered radio access network
%         \item \emph{user} - handled user
%     \end{itemize}
%     \textbf{Result: } \emph{user} served by one of \emph{cells}
%     \begin{algorithmic}[1]
%         \State calculate decision criteria values for each radio link between the \emph{user} and each cell from list \emph{cells}: current load of the cell, percentage of cell's available radio resources that would be used up upon serving the \emph{user} and an estimated remaining number of users to be handled by the cell
%         \State \emph{criteria} $\leftarrow$ results of the decision criteria calculation
%         \State input \emph{criteria} into the neural network's first layer
%         \State \emph{data} $\leftarrow$ neural network's last layer output
%         \State calculate non-viable connections filtering matrix
%         \State \emph{filter} $\leftarrow$ result of filtering matrix calculation
%         \State \emph{filtered data} $\leftarrow$ \emph{filter} \(\times\) \emph{data}\
%         \State \emph{index} $\leftarrow$ \emph{argmax}(\emph{filtered data})\
%         \State \emph{target cell} $\leftarrow$ \emph{cells}[\emph{index}]\
%         \State allocate \emph{target cell}'s radio resources to the \emph{user} 
%     \end{algorithmic}
% \end{algorithm}

\section{Scenario description}
\subsection{Base stations and user distribution}
Simulation scenario for (downlink) traffic steering algorithms' efficiency comparison includes an urban environment with five base stations, including both LTE-A and NR radio access technologies. In the center, there is an LTE-A macro cell which can transmit with a maximum power of 43 dBm; two LTE-A and two NR micro cells are deployed within its coverage range, creating the HetNet scenario. The transmit power of the former cells is 32 dBm, whereas for the latter -- 34 dBm. The operating frequencies for the LTE-A and NR cells are 2100 MHz and 3500 MHz, respectively. 
%The key configuration parameters of the base stations configuration are listed in Table \ref{tab:bs_list}

% \begin{table}[!t]
% \renewcommand{\arraystretch}{1.3}
% \caption{Base station parameters in simulation scenario}
% \label{tab:bs_list}
% \centering
% \begin{tabular}{|c|c|c|c|c|c|c}
%      \hline
%      Nr & RAT & Cell size & X pos. & Y pos. & TX power \\
%      \hline
%      1. & LTE & macro & 1750 m & 1750 m & 43 dBm \\
%      \hline
%      2. & LTE & micro & 1400 m & 1050 m & 32 dBm \\
%      \hline
%      3. & LTE & micro & 1400 m & 2275 m & 32 dBm \\
%      \hline
%      4. & NR & micro & 2800 m & 1155 m & 34 dBm \\
%      \hline
%      5. & NR & micro & 2800 m & 1155 m & 34 dBm \\
%      \hline
% \end{tabular}
% \end{table}

\par
Users are distributed randomly in range of each base station, i.e. 300 users in range of the macro cell and 60 users in range of each micro cell. Each user is assigned one of the three available user profiles (as defined in Tab.~\ref{tab:u_profiles}) with specific probabilities.
%Most of the simulated users are voice users with demanded bit rate equal to 96 kbps. Remaining two groups of users are data-heavy users with two levels of demanded bit rate: 5 Mbps (adequate to 720p 30fps video streaming) and 24 Mbps (adequate to 1440p 60fps video streaming). %\cite{yt_bitrates}.
\begin{table}[!htb]
\renewcommand{\arraystretch}{1.1}
\caption{User profile parameters in simulation scenario}
\label{tab:u_profiles}
\centering
\begin{tabular}{|c|c|c|c|c|}
     \hline
     Profile name & Probability & Bit rate demand \\
     \hline
     Voice (low rate) & 75\% & 96 kbps \\
     \hline
     Data (mid rate) & 20\% & 5 Mbps \\
     \hline
     Data (high rate) & 5\% & 24 Mbps \\
     \hline
\end{tabular}
\end{table}
%In terms of radio link simulation, only downlink is considered for each user. Handling order of the users is random and different for each simulation episode\footnote{A simulation episode is considered as a process of allocating radio resources to all of the users.}. For each traffic steering method 30000 simulation episodes are considered.

\subsection{Radio channel quality model}
%In a given user placement radio channel quality must be determined in order to assess user satisfaction with chosen radio resource allocation and fairly compare considered traffic steering algorithms' efficiency.
Standardized channel model, described in \cite{etsi_prop_model}, was used to calculate line of sight (LOS) probabilities and pathloss for each user-cell pair. The model has been utilized for both LTE-A and NR cells, as it is declared viable for frequencies from 500 MHz to 100 GHz. As the considered simulation scenario includes an urban environment, UMa and UMi variants of the channel model are used for macro and micro cells, respectively.
\par
Based on the calculated pathloss, transmitted power of the given cell and gains/losses related to base station's and user's equipment, a value of signal power received by the user is determined for each base station assuming the receiver noise sensitivity at -110 dBm. The SINR value is then compared with target values in the standardized CQI (e.g. \cite{nr_cqi}) to find an estimated code rate used for further effective bit rate calculations. In the simulation, 20 MHz bandwidth was considered with 15 kHz and 30 kHz of subcarrier spacing for LTE-A and NR,~accordingly. 

\subsection{Traffic steering process}
Each simulation episode includes an allocation of radio resources to each user. At the beginning of the episode, it is assumed that all base stations have all radio resources available, and that no user is yet served by the network. The handling order of users is random. When the resources of all base stations in the test scenario are exhausted, the episode ends. It is assumed that the allocation of resources takes place within the network controller which has information about the load of each base station, the parameters of the signals received by the user's equipment and the bitrate demand.

\section{Simulation results}
To evaluate the efficiency of the proposed solution extensive computer simulations have been carried out. Per each method 30000 episodes have been considered to guarantee statistical reliability of the results. %Efficiency measurements of the traffic steering methods were performed during the simulation of algorithms with a random distribution of users, as shown in Figure \ref{fig:simulation_scenario}. Simulation of each traffic steering method consisted of 30,000 episodes. Such a number of episodes was selected in order to achieve a compromise between the computation time and the minimization of errors in the averaged values of the efficiency statistics.
%\par
Each method's efficiency was measured against two statistics collected at the end of each episode: mean user satisfaction (MUS), and mean not-handled-user (NHU) count. The former one ranges from 0 to 1 and is calculated as the mean ratio of demanded to received bitrate among all users. The number of NHUs is determined at the end of the episode as the number of users with a satisfaction of 0.
\par
Efficiency statistics for CLB, SLB and RLLB method are listed in Table \ref{tab:results}, where \(S_{av}\) is the MUS after 30000 simulation episodes with \(\sigma_S\) being the standard deviation of mean satisfaction values and \(N_{av}\) is the mean NHU count after 30000 simulation episodes with \(\sigma_N\) being the standard deviation of mean NHU counts.
\begin{table}[!htb]
\renewcommand{\arraystretch}{1.1}
\caption{Efficiency statistics results for considered traffic steering methods in the test scenario}
\label{tab:results}
\centering
\begin{tabular}{|c|c|c|c|}
     \hline
     Method & CLB & SLB & RLLB \\
     \hline
     \(S_{av}\) & 0.860 & 1.000 & 0.998 \\
     \hline
     \(\sigma_S\) & 0.020 & 0.000 & 0.002 \\
     \hline
     \(N_{av}\) & 74.80 & 75.98 & 73.89 \\
     \hline
     \(\sigma_N\) & 11.02 & 10.94 & 10.87 \\
     \hline
\end{tabular}
\end{table}

The CLB method provides the lowest MUS of 86\%, with its standard deviation equal to 2\%. The SLB method managed to achieve 100\% MUS in every episode. The RLLB method is almost as efficient in this respect as the SLB method; it managed to achieve near 100\% MUS, with its standard deviation ten times lower than in the CLB method. Next, in terms of mean NHU count, the SLB method is the least efficient, as it leaves almost 76 users not handled on average. The CLB method is capable of handling one user more on average. The most efficient method in this respect is the RLLB method, which manages to handle two users more than the SLB method on average. %Standard deviation of the mean NHU counts for each method is similar.
Moreover, the RLLB method manages to combine the best features of both reference methods, with almost 100\% MUS as the SLB method and a relatively small NHU count as the CLB method. %It can therefore be concluded that the RLLB method offers near ideal MUS and at the same time allows to serve the largest number of users with the same resources.

\section{Conclusion}
The novel algorithm called Reinforcement Learning Load Balancing successfully uses an ANN trained with the SARSA algorithm for optimal traffic control in radio access networks. The method adjusts its decisions based on feedback from the network in form of user satisfaction observed during the simulation episode. Decisions made with use of the continuously optimized ANN model allow for achievement of a small number of NHUs and high satisfaction of the users served.
\printbibliography
%\begin{thebibliography}{1}
%\bibitem{IEEEhowto:kopka}
%H.~Kopka and P.~W. Daly, \emph{A Guide to \LaTeX}, 3rd~ed.\hskip 1em plus
%  0.5em minus 0.4em\relax Harlow, England: Addison-Wesley, 1999.
%\end{thebibliography}
% that's all folks
\end{document}